\documentclass{article}

\usepackage{nac_preprint}

\usepackage{times}
\usepackage{soul}
\usepackage{url}
\usepackage[hidelinks]{hyperref}
\usepackage[utf8]{inputenc}
\usepackage[small]{caption}
\usepackage{graphicx}
\usepackage{subcaption}
\usepackage{booktabs}
\usepackage{multirow}
\usepackage{algorithm}
\usepackage{algorithmic}
\usepackage[numbers]{natbib}

\usepackage{amsmath}
\usepackage{amsfonts}
\usepackage{amssymb}
\usepackage{mathtools}
\usepackage{amsthm}

\usepackage[capitalize,noabbrev]{cleveref}

\theoremstyle{plain}
\newtheorem{theorem}{Theorem}[section]
\newtheorem{proposition}[theorem]{Proposition}

\theoremstyle{definition}
\newtheorem{definition}[theorem]{Definition}

\theoremstyle{remark}

\usepackage[textsize=tiny]{todonotes}

\title{Faster Predictive Coding Networks \\ via Better Initialization}

\author{
    \textbf{Luca Pinchetti$^{1}$, Simon Frieder$^{1}$, Thomas Lukasiewicz$^{1,2}$, Tommaso Salvatori$^{2,3}$}
    \\ $^1$Department of Computer Science, University of Oxford, Oxford, UK
    \\ $^2$Institute of Logic and Computation, Vienna University of Technology, Vienna, Austria
    \\ $^3$VERSES AI Research Lab, Los Angeles, US
    \\
    \\ \texttt{luca.pinchetti@cs.ox.ac.uk}
}

\begin{document}

\maketitle

\begin{abstract}
Research aimed at scaling up neuroscience inspired learning algorithms for neural networks is accelerating. Recently, a key research area has been the study of energy-based learning algorithms such as predictive coding, due to their versatility and mathematical grounding. However, the applicability of such methods is held back by the large computational requirements caused by their iterative nature. In this work, we address this problem by showing that the choice of initialization of the neurons in a predictive coding network matters significantly and can notably reduce the required training times. Consequently, we propose a new initialization technique for predictive coding networks that aims to preserve the iterative progress made on previous training samples. Our approach suggests a promising path toward reconciling the disparities between predictive coding and backpropagation in terms of computational efficiency and final performance. In fact, our experiments demonstrate substantial improvements in convergence speed and final test loss in both supervised and unsupervised settings.
\end{abstract}

\section{Introduction}
In the last decade, neural networks trained with backpropagation (BP) have surpassed human-level performance in a multitude of specialized tasks, such as natural language processing \cite{chen2020big} and image recognition \cite{he2016deep} and generation \cite{ramesh2022hierarchical}. However, recent improvements are to be mostly attributed to the exponential scaling of such models rather than to major shifts in the underlying technology \cite{GPT3, achiam2023gpt}. The computational cost to train such models is becoming prohibitive, and alternatives must be sought. Biologically plausible learning algorithms are parallel and local, key properties that make them especially well suited for analog and neuromorphic hardware \cite{hagiwara2024design,scellier2023energy}, which should provide faster and more energy-efficient machines. Among them, \textit{predictive coding} (PC) \cite{rao1999predictive} has been shown to be a flexible algorithm \cite{salvatori2022learning} that can be used to train neural networks for multiple tasks \cite{pinchetti2022beyond,zahid2023sample, salvatori2021associative}.

PC networks are trained through a variation of generalized expectation maximization (EM) \cite{friston2008hierarchical}, according to which each layer (i.e., group of neurons in the network) is first independently updated to minimize an objective known as the energy of the network (expectation phase) and, successively, the weights of the network are modified to further reduce it (maximization phase). The two steps are commonly referred to as \textit{inference} and \textit{learning} \cite{whittington2017approximation}. As inference requires multiple updates to reach convergence, it introduces a non-negligible computational cost that has been hindering the applicability of biologically plausible learning algorithms to practical real-world tasks on existing GPU hardware. In fact, while all its operations are strictly local and can hence run in parallel, the inference phase must run for at least $L$ steps in order to produce meaningful results, where $L$ is the number of layers of the neural network. This is to make sure that input label information propagates towards the whole network, and hence all the parameters of the model are updated accordingly. 

Recent work in the field has mostly focused on improving such models in terms of test accuracy \cite{pinchetti2024benchmarking, qi2025towards, goemaere2025error}. To this end, a major challenge was to solve the stability and performance in models with more than $10$ layers. However, despite the progress achieved in this direction, both in terms of theoretical understanding and performance  \cite{innocenti2025mu, qi2025towards}, little work has been done to speed these models up. In this work, we tackle the efficiency problem by quickly inferring better new neuron initialization methods, that allows a single training iteration to be completed in less than $L$ inference steps. Our contributions can be summarized as follows:
\begin{itemize}
    \item We analyze and compare different existing neuron initialization techniques for PC networks, providing the first comprehensive discussion on the topic. By doing so, we highlight current limitations and shortcomings of each method. Our goal is to underline the importance of finding a good initialization for PC networks.
    \item We propose a new intuitive initialization technique, which we call \emph{stream-aligned average initialization}, and formally prove on a toy network that it is an improvement over the existing best method. We then generalize it to unsupervised learning settings, by enhancing PC layers with continuous Hopfield networks (HNs) \cite{ramsauer2020hopfield} and using their associative memory capabilities to initialize the network.
    \item We evaluate our method on common PC tasks, empirically showing that it performs better than standard PC in terms of test accuracy, with lower training times (up to 5 times faster in terms of sequential matrix multiplications, a metric used to measure energy-based networks complexity \cite{salvatori2022incremental,alonso2023understanding} in all the considered supervised-learning benchmarks, and providing an even better performance in unsupervised tasks. These improvements allow PC to train neural networks with a computational efficiency comparable to backpropagation, finally bridging the gap between neuro-biologically inspired and standard deep learning.
\end{itemize}


\label{introduction}

\section{Predictive Coding}
\label{predictive-coding}
\begin{figure}[t]
    \centering
    \includegraphics[width=0.6\columnwidth]{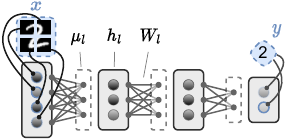}
    \caption{Discriminative PCN with $4$ layers. For supervised training, we fix the first and last layers to the input-output pair $(x,y)$.}
    \label{fig:pcn}
\end{figure}

Predictive coding networks (PCNs) are energy-based models, meaning that a solution is reached by minimizing an energy function defined on all the parameters of the neural network. Structure-wise, they are similar to BP-trained deep networks, with neurons organized in $L + 1$ layers $(h_0,\dots,h_L)$ connected by $L$ weight matrices $(W_0,\dots,W_{L-1})$. During (supervised) training, the input and output neurons, $h_0$ and $h_{L}$, are fixed to the input-output pair $(x,y)$, meaning that their value will not change during inference, while the remaining neurons and the weights are updated via gradient descent to minimize the energy $\mathcal{F}$, defined as follows:
\begin{equation}
    \mathcal{F}(\mathbf{h}, \mathbf{W})  = \frac{1}{2}\sum\nolimits_{i=1}^{L} ||h_l - \mu_l||^2,
\end{equation}
where $\mathbf{h} = (h_0, ..., h_{L})$, $\mathbf{W} = (W_0, ..., W_{L-1})$, and $\mu_l$ is the prediction produced by layer $l-1$:
\begin{equation}
    \mu_l = f_{l-1}(h_{l-1}, W_{l-1}),
\end{equation}
and $f_{l-1}$ is any standard layer transformation such as fully connected (in which case, $f_{l-1}(h_{l-1}, W_{l-1}) = \sigma(\hat{W}_l h_{l-1} + \hat{b}_l)$, with $W_{l-1} = (\hat{W}_l,\hat{b}_l)$) or convolutional. The difference between the predicted and actual value of every layer, i.e.,  $e_l = h_l - \mu_l$, is called prediction error, hence the name predictive coding. Fig.~\ref{fig:pcn} provides an example of a PCN.

As formally highlighted by \cite{frieder22non-convergence}, the inference updates generate a discrete dynamical system over the state of the network neurons $\mathbf{h}$, governed by the gradient update equation:
\begin{equation}
\label{eq:delta-x}
    \mathbf{h}^{(t+1)} = \mathbf{h}^{(t)} - \alpha \frac{\partial \mathcal{F}(\mathbf{h}^{(t)}, \mathbf{W})}{\partial h^{(t)}} 
\end{equation}
which per-layer equals 
\begin{equation*}
    h_{l+1}^{(t+1)}= h_l^{(t)} - \alpha (\frac{\partial f_l(h_l^{(t)}, W_l)}{\partial h_l^{(t)}}e_{l+1}^{(t)} - e_l^{(t)}),
\end{equation*}
for every $l=0,\ldots,L$,
where $\alpha$ is the neurons learning rate. Typically, to simulate convergence of the expectation step, inference is run for a number of steps $T$, which is linearly proportional to the amount of mappings $L$ ($T \approx 5L$ seems to be an average value used \cite{whittington2017approximation, salvatori2022learning, pinchetti2022beyond}). After $T$ inference steps, learning is obtained by updating the network weights as follows:
\begin{equation}
\label{eq:delta-w}
\begin{split}
    W_l^{(b+1)} &= W_l^{(b)} - \beta \frac{\partial \mathcal{F}(\mathbf{h}^{(T, b)}, \mathbf{W}^{(b)})}{\partial W_l^{(b)}} = W_l^{(b)} - \beta (e_{l+1}^{(t)}f_l(h_l^{(T, b)}, W_l^{(b)})^T),
\end{split}
\end{equation}
where $\beta$ is the weights learning rate, and $b$ indexes the current mini-batch given to the model during training. Eqs.~\eqref{eq:delta-x} and~\eqref{eq:delta-w} explicate the local nature of PC, as every update depends exclusively on pre- and post-synaptic values and is, therefore, Hebbian \cite{hebb2005organization}. It follows that all layer and weight updates can be executed concurrently  during the inference and learning phases, respectively. To theoretically measure the efficiency of such algorithm, \cite{salvatori2022incremental} introduced a metric, called sequential matrix multiplications (SMMs), used to estimate the time requirements of a weight update performed via PC, when assuming a complete parallelization of every layer computation. Intuitively, this metric assumes that the number of sequential matrix multiplications is a fair approximation of the efficiency of an algorithm, as matrix multiplications are usually (by far) the most expensive operation. They concluded that PC and BP require
\begin{equation}
\label{eq:SMM_PC}
    \operatorname{SMM}_{PC}=2T \ \ \ \ \ \ \text{and} \ \ \ \ \ \ \operatorname{SMM}_{BP} = 2L-1,
\end{equation}
SMMs per weight update, respectively. Considering the average used $T$, we see how the complexity of PC currently far exceeds that of BP\footnote{Especially when factoring in that current hardware and deep learning libraries do not seem to offer efficient parallelization of independent computational graphs \cite{JiaRu2016_2021, 2021xla}, making the computational cost of training a PCN quadratic over $L$.}.
\section{Existing Initialization Techniques}
\label{initialization}
The attention towards PC from the machine learning community is quite recent \cite{bogacz2017tutorial, whittington2017approximation, song2020can}. This paradigm shift from neuroscience (where the goal is to study biological learning) to the machine learning world (where the focus is on performance) has created the need of adapting PC to enhance its applicability to practical tasks. To this end, to make PC more efficient, we are limited to using a relatively small number of inference steps $T$: this makes inference much faster, but is inadequate for the convergence of the energy function $\mathcal{F}$. We claim that a key factor that influences convergence speed is the initialization technique used for the neurons of the network. As there exists no analysis studying the initialization of a PCN, and different works seem to opt for different solutions without explicating their choice or comparing different options, in this section, we propose a formal study and comparison of such methods.\footnote{To simplify the notation, in this section, we avoid specifying the batch index $b$ (i.e., $\mathbf{h}^{(0, b)}$) when not necessary, as introduced in Section \ref{predictive-coding}. The full notation used is explained in Table~\ref{tab:notation}.}  
Let $\mathbf{h}^{(0)}$ be the state of the neurons of a PCN at inference step $t=0$. We define an initialization $\mathcal{I}$ as an assignment of $\mathbf{h}^{(0)} = (h_l^{(0)})_{l \in \{0,...,L\}}$. Note that, depending on the settings, $h_0$ and/or $h_{L}$, may be clamped to some given input/output data (i.e., $h_0$ and/or $h_{L}$, may be given fixed values or allowed to vary, depending on whether we do inference or training). We identify the following methods:
\begin{enumerate}
    \item \textbf{Random} ($\mathcal{I}_{\mathcal{N}}$). Each layer initial value is sampled from a distribution $\mathcal{D}$ with constant sufficient statistic $\phi$: $h^{(0)}_l \sim \mathcal{D}(\phi),\;\forall l \in \{0, ..., L\}$. Commonly, $\mathcal{D}(\phi) = \mathcal{N}(a,b)$, with a normal distribution $\mathcal{N}$ with mean $a$ and variance $b$ \cite{salvatori2022learning};
    \item \textbf{Zero}\, ($\mathcal{I}_{0}$). $h^{(0)}_l = 0,\;\forall l \in \{0, ..., L\}$. We can consider $\mathcal{I}_{0}$ as a special case of $\mathcal{I}_{\mathcal{N}}$, $a = 0$ and $b \to 0^+$. This is used, for example, by \cite{ororbia2022neural};
    \item \textbf{Null} ($\mathcal{I}_{\varnothing}$). The neurons are not re-initialized after a weight update, but their value is instead preserved: $h^{(0, b+1)}_l = h^{(T, b)}_l$. This technique has shown promising results in \cite{salvatori2022incremental}, which, however, focuses on full-batch training. With minibatches, the existing neuron values will, instead, contain information unrelated to the new samples provided after a weight update and, thus, produce noisy energy values;
    \item \textbf{Forward} ($\mathcal{I}_{\text{fw}}$). The neurons are initialized from input to output by forwarding the activation of each layer: $h_l^{(0)} = \mu_l^{(0)},\;\forall l \in \{0, ..., L\}$ \cite{whittington2017approximation, pinchetti2022beyond, alonso2023understanding}%
    \footnote{For completeness, we also report the existence of what can be seen as a \textit{reverse} forward initialization, which is used to help with \textit{backward} inference (i.e., predicting the input from the output) on classification tasks \cite{tscshantz2023hybrid}. However, in this work, we focus on \textit{forward} networks, since we want to establish a direct comparison with backpropagation. Furthermore, the method requires learning a mirrored set of new weights for each layer and does not fully address the limitations of $\mathcal{I}_{\text{fw}}$.}.
\end{enumerate}
\begin{figure}[t]
    \begin{center}
    \centerline{\includegraphics[width=0.6\columnwidth]{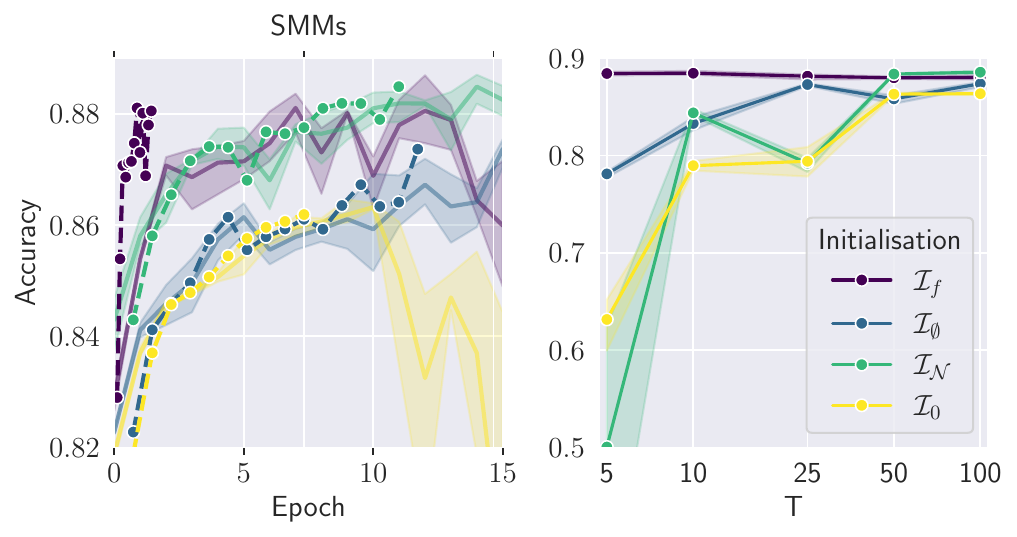}}
    \caption{Performance of different initialization methods. (Left) Training curves with test accuracy over epochs (solid lines) and SMMs (dotted lines, plotted until the highest accuracy is reached for each method). The shaded area corresponds to the standard deviation over 3 seeds. (Right) Impact of the number of inference steps $T$. Only $\mathcal{I}_{\text{fw}}$ is able to train the model at low values of $T$.}
    \label{img:init_summary}
    \end{center}
\end{figure}
\begin{figure*}[ht]
    \centering
    \includegraphics[width=1\columnwidth]{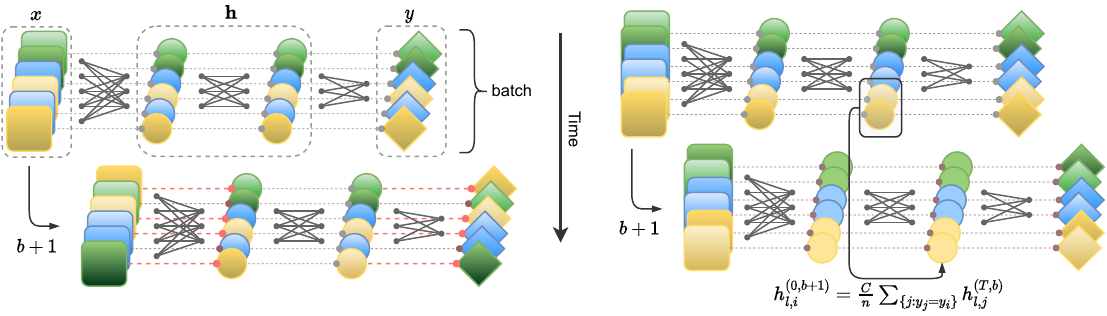}
    \caption{Comparison between batch training and stream aligned training on a classification task. Different colors represent different classes. (Left) \textbf{Batch training:} samples are randomly shuffled within a batch; consequently the hidden state of the network after $T$ steps cannot be preserved, as new samples may belong to different classes, which would result in high error values in the first and last layers. (Right) \textbf{Stream-aligned training:} the hidden state is always associated with elements of the same class, reducing the data distribution variance and, consequently, the error. Furthermore, by taking class-wise averages of the hidden neurons, we further spread the energy through the network, as highlighted by the red-shaded lines (saturation represents error).}
    \label{fig:training_comparison}
\end{figure*}
Note that such techniques may even be combined: for example, in the original paper, null initialization $\mathcal{I}_{\varnothing}$ can only be applied from the second weight update onward, and the network is first initialized via forward initialization \cite{salvatori2022incremental}.
In classification tasks, $\mathcal{I}_{\text{fw}}$ is the most used initialization due to its interesting properties. It was shown that it guarantees perfect layer initialization during evaluation, as it sets the neurons to the fixed point determined by Eq.~\eqref{eq:delta-x}, satisfying $\mathcal{F}(\mathbf{h}^{(0)}, \mathbf{W}) = 0$ \cite{frieder22non-convergence}. Furthermore, during training, the energy is concentrated in the last layer, since $e_{l \neq L+1} = 0$ and $\mathcal{F}(\mathbf{h}^{(0)}, \mathbf{W}) = \frac{1}{2}(\mu_{L}^{(0)} - y)^2$, where $y$ is the output label to which $h_{L}$ is fixed to. This makes the network initialized in such a way that, at $t=0$, the energy function is equivalent to the loss function of BP \cite{millidge2022theoretical}.

\textbf{Experimental comparison:} we analyze the learning speed of applying the above listed methods to a 5-layers fully connected network for FashionMNIST classification \cite{FashionMNIST}. Details are given in the supplementary material. Our aim is to show that initialization plays a significant role in PCN performance. Since computational complexity is proportional to the number of iterative inference steps, we compare behaviors for different $T$ values, and report them in Fig.~\ref{img:init_summary}. The results show that (i) as expected, $\mathcal{I}_{\text{fw}}$ is the best performing initialization, able to achieve a good performance at the lowest $T$ values; (ii) $\mathcal{I}_{\varnothing}$ outperforms both $\mathcal{I}_0$ and $\mathcal{I}_\mathcal{N}$, suggesting that hidden neurons for samples in consecutive minibatches are, at least partially, correlated and preserving the PCN hidden state provides better results than random initialization; and (iii), that all initialization methods reach similar accuracy values only at $T=100$, showing that different initialization play an important role mainly when we need faster training times, but less if have the time allow the model to run for many iterations. 

\subsection{Limitations of forward initialization}
\label{limitations}
We now discuss two limitations of the forward initialization $\mathcal{I}_{\text{fw}}$: its low efficiency, and its inability to generalize to models with cyclic structures. In the first case, establishing an analytical lower bound $T^{\min}$ for the number of inference steps required to train a PCN is currently an open problem. However, since $\mathcal{I}_{\text{fw}}$ concentrates all the energy in the output layer, at least $T^{\min}_f = L$ steps are necessary to propagate the error through the network, back to the input neurons. Consequently, even under hypothetically optimal conditions, by substituting $T^{\min}_f$ into Eq.~\eqref{eq:SMM_PC}, we have that $\operatorname{SMM}_{\mathcal{I}_{\text{fw}}} \geq 2L$. Furthermore, we note that $\mathcal{I}_{\text{fw}}$ requires a forward sweep to initialize the hidden layers, which requires $L$ additional sequential multiplications (one for each layer). This brings the total SMM complexity for a single batch of datapoints to
\begin{equation}
\label{eq:SMM_I_f}
     \operatorname{SMM}_{\mathcal{I}_{\text{fw}}} \leq 2L + L = 3L,
\end{equation}
which is still higher than $\operatorname{SMM}_{BP}$. It is also unclear how to extend forward initialization to complex PCN structures, such as the ones shown in \cite{salvatori2022learning}, which may include cyclic connections or be generative models trained only by providing the output (i.e., by fixing the output neurons). Here, we propose a formal definition for $\mathcal{I}_{\text{fw}}$ and show that this initialization is only meaningful if the architecture of a PCN forms a directed acyclic graph (DAG) on the level of neurons. This corresponds to the class of models of classical deep neural networks trained via BP, which, however, is only a subset of possible PCNs.
\begin{definition}
\label{def:extended_forward}
Forward initialization $\mathcal{I}_{\text{fw}}$ can be trivially extended to a PCN defined over any directed acyclic graph, where a set of input neurons is designated, in the following way: Analogously to the case for layered architectures, first input values are associated to the input neurons. Then, we successively follow each neuron child through the graph and initialize it by the sum of all incoming predictions:
\begin{equation*}
    \mu_l = \sigma(\sum_{i \in Parents(l)} f_{i}(h_{i}, W_{i})).
\end{equation*}
\end{definition}
\begin{proposition}
\label{thm:forward_dag}
If forward initialization of a PCN $\mathcal{M}$, as defined above, is possible, then $\mathcal{M}$ is defined on a DAG $\mathcal{G}$. Furthermore, all root nodes of $\mathcal{G}$ must be fixed to known inputs.
\end{proposition}
A sketch of the proof of Proposition \ref{thm:forward_dag} is given in Appendix~\ref{thm:forward_dag}.
As forward initialization makes the internal energy of the model equal to zero, and thus minimize it, note that a more general definition, which would extend to the case of cyclic structures, is that of equilibrium propagation \cite{scellier2017equilibrium}, where the energy is minimized twice: the first time, with no label information, and the second time by fixing the output neurons to the label vector. However, such a method does not address the efficiency problem on GPUs, as it doubles the number of iterations needed to perform a weight update. 
\section{Methods}
\label{methods}
In Section \ref{limitations}, we proved that $\mathcal{I}_{\text{fw}}$ is only meaningful when dealing with a specific subclass of PCNs. Here, we intuitively suggest that it is sub-optimal also during training and that, consequently, existing state-of-the-art PC performance can be significantly improved upon by introducing better initialization techniques. We argue that ``resetting" the neurons with a forward pass when receiving a new minibatch $b+1$ can be detrimental, as information about convergence values for batch $b$ is lost in the process and the network needs more inference steps than otherwise. In fact, after a weight update $\mathbf{W}^{(b+1)} = \mathbf{W}^{(b)} + \beta \Delta\mathbf{W}^{(b)}$, the hidden state convergence point will have only slightly changed and, thus, $\mathbf{h}^{(T, b)}$ is still a good approximation of the new convergence point $\mathbf{h}^{(T, b+1)}$, assuming the samples of minibatch $b+1$ are similar enough to the ones in $b$. This also explains why the method proposed in \cite{salvatori2022incremental} is effective under full-batch settings as each batch sample $i$ is fixed throughout training.
%
%

\subsection{Stream-aligned training and average initialization}
Under most circumstances, however, minibatch training is the only feasible option, and the training samples are drawn from a complex distribution, which introduces a high variance between data points in consecutive batches. Consequently, $\mathbf{h}^{(T, b)}$ is not a good approximator for $\mathbf{h}^{(T, b+1)}$ anymore, as shown in  Fig.~\ref{fig:training_comparison}. In order to reduce the sample variance and re-use the hidden state $\mathbf{h}^{(T)}$, we suggest a new training setting specific for classification tasks, which we call \textit{stream-aligned training with average initialization}. Intuitively, instead of having data samples randomly shuffled within a minibatch, we ``align" them based on their label, effectively sampling from $C$ distributions $\mathcal{D}_c$ with low sample variance, where $C$ is the number of classes. We divide a minibatch $b$ of size $n$ in $C$ sub-minibatches $b_c$ of size $\frac{n}{C}$, each defined as follows:
\begin{equation*}
    [(x_i, y_i = c)]_{i\in (1, ..., \frac{n}{C})},\; x \sim \mathcal{D}_c,
\end{equation*}
where index $i$ identifies a sample in a minibatch $b$: $\mathbf{h}_i = (h_{0,i}, ..., h_{l,i}, ..., h_{L,i})$ is the neurons state of sample $i$. To further reduce the energy variance between samples of the same class, we average the neuron values of each ``stream" over their respective class (Fig.~\ref{fig:training_comparison}):
\begin{equation}
\label{eq:avg_init}
    h^{(0, b+1)}_{l,i} = \frac{C}{n} \sum\nolimits_{\{j: y_j = y_i\}} h^{(T, b)}_{l, j}.
\end{equation}
We refer to this initialization as $\mathcal{I}_{\text{avg}}$. Note that the computations described in Eq.~\eqref{eq:avg_init} are layer-wise local and do not involve matrix multiplications. Therefore, $\mathcal{I}_{\text{avg}}$ does not add to the SMM complexity of training a PCN, contrarily to $\mathcal{I}_{\text{fw}}$ (Eq.~\eqref{eq:SMM_I_f}).

$\mathcal{I}_{\text{avg}}$ can be framed as modelling an approximated converged hidden state distribution conditioned on the label $y$ and current model weights $\mathbf{W}$, $p_{\mathbf{W}}(\mathbf{h}^{(T)}|y)$. Importantly, we are not learning it as a single Markov chain $p_{\mathbf{W}}(h_{l-1}^{(T)}|h_{l}^{(T)}),\;\forall l \in \{1, ..., L\}$, but as many conditional distributions $p_{\mathbf{W}}(h_l^{(T)}|y)$. Note that, as this initialization depends on the label, this can only be considered for training, where we have access to label information. During inference, we will again use $\mathcal{I}_{\text{fw}}$ and forward the input $x$ through the layers, guarantying $\mathcal{F} = 0$ \cite{frieder22non-convergence}. While testing and training using different methods is unusual in machine learning, we will empirically show the better performance of such technique. Ultimately, $\mathcal{I}_{\text{avg}}$ initializes the network in a better state during training, adding no overhead during evaluation.

In practice, as we will discuss in Section \ref{experiments}, a hybrid approach between $\mathcal{I}_{\text{fw}}$, used to initialize the first layers, and $\mathcal{I}_{\text{avg}}$, used for the remaining ones, seems to work the best. $\mathcal{I}_{\text{avg}}$ has the advantage of requiring no extra model parameters and only lightweight computations. 
For the sake of clarity, in this section, we assumed a balanced training dataset with a limited number of classes, which justifies the use of sub-minibatches with equal and fixed size such that $\frac{C}{n} \leq 1$. However, as we will show in the next section, it is possible to generalize beyond these constraints using a \textit{memory-based} initialization.

\begin{figure}[t]
    \centering
    \includegraphics[width=0.6\columnwidth]{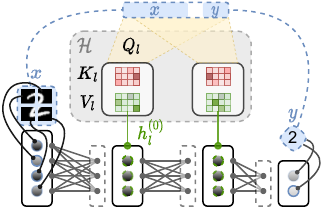}
    \caption{PCN whose layers are enhanced with associative memories for better initialization.}
    \label{fig:pcn_hn}
\end{figure}
\subsection{Unsupervised learning and memory-based initialization}
A major drawback of stream-aligned training with average initialization is the dependence on classification labels, which makes this approach not generalizable to regression tasks or unsupervised learning. To address this, we consider input-label pairs $(x, y)$ separately, we here denote observations (data points) as  $o = (x, y)$, and enhance the layers of a PCN with an external memory $\mathcal{H}$. Such memory, which comes at the cost of a few extra model parameters, is then used to approximate the full conditional distribution $p_{\mathbf{W}}(\mathbf{h}^{(T)}|o)$. 

To store and retrieve memories, we use modern Hopfield networks \cite{ramsauer2020hopfield}, a continuous version of Hopfield networks \cite{hopfield1982neural} with exponential memory capacity. We use \textit{HopfieldLayers} (i.e., layers that contain a value matrix storing the memorized patterns, as well as query and key matrices to retrieve them) as associative memory, which results in the following initialization:
\begin{equation}
\label{eq:mem_init}
    h^{(0)}_{l,i} = \sigma(\delta_l (o_i Q_l + b_{l}) K_l^T) V_l,
\end{equation}
where $\sigma$ is the \textit{softmax} function, $o_i$ is the concatenation of all observations for sample $i$, and $Q_l$, $K_l$, and $V_l$ are the projection matrices of the attention mechanism called \emph{query}, \emph{key}, and \emph{value} \cite{vaswani2017attention}. Furthermore,  $b_l$ is a learnable bias, and $\delta_l$ is the inverse temperature. We refer to this initialization as $\mathcal{I}_{\text{mem}}$. The obtained PCN is presented in Fig.~\ref{fig:pcn_hn}. To train $\mathcal{H}$, we simply use the squared-error loss:
\begin{equation}
\label{eq:mem_loss}
    \mathcal{L}_\mathcal{H} = \sum_{l=1}^L \sum_{i=1}^n (\sigma(\delta_l (o_i Q_l + b_{l}) K_l^T) V_l - h^{(T)}_{l,i})^2.
\end{equation}
and minimise it via gradient descent at each mini-batch, treating $\mathcal{H}$ analogously to other model parameters. Before training, we initialize $V_l$ to minimize the difference between $\mathcal{I}_{\text{mem}}$ and $\mathcal{I}_{\text{fw}}$, such that both memory and untrained network share a similar starting point. To do so, we set $V_l = A_l^+ h^{(0)}_{\mathcal{I}_{\text{fw}},l}$, where $A_l^+$ is the pseudo-inverse of $A_l = \sigma(\delta_l (O Q_l + b_{l}) K_l^T)$, with $O$ being the known observations matrix of the first minibatch, and $h^{(0)}_{\mathcal{I}_{\text{fw}},l}$ is the state of hidden layer $h_l$ initialized with $\mathcal{I}_{\text{fw}}$.

\section{Experiments and Results}
\label{experiments}
We compare the results of deep neural networks on various tasks, comparing BP with PC and analyzing the effectiveness of the two initialization methods proposed in this work. In supervised learning settings, PC-$\mathcal{I}_{\text{avg}}$ significantly outperform the current state-of-the-art $\mathcal{I}_{\text{fw}}$ in terms of both accuracy and training speed. Furthermore, it also overtakes BP on fully connected networks and closely matches it on the more complex architecture considered in this work, AlexNet \cite{krizhevsky2012imagenet}.
\begin{figure}[ht]
    \centering
    \includegraphics[width=0.3\columnwidth]{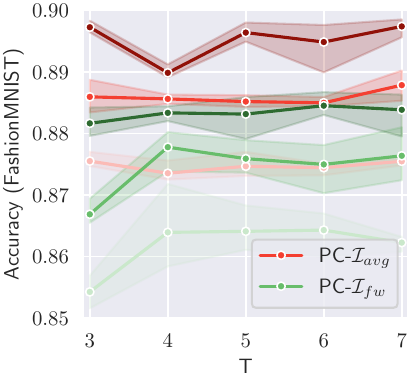}%
    \hspace{1em}%
    \includegraphics[width=0.3\columnwidth]{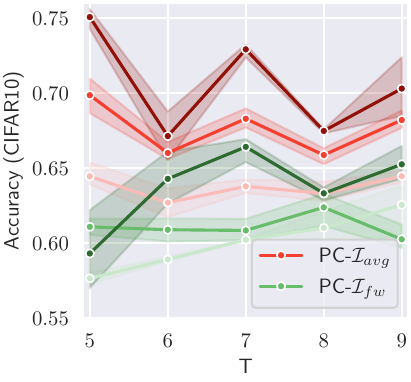}
    \caption{Test accuracy for different values of inference steps $T$ used during training. Shades
of color indicates different fractions of training data used (i.e., $25\%$ is lighter; $50\%$, $100\%$ are darker).}
    \label{fig:accuracy_over_T}
\end{figure}
\begin{table*}[ht]
\centering
\vspace{-1em}
\caption{Test accuracy of BP, PC-$\mathcal{I}_{\text{fw}}$, PC-$\mathcal{I}_{\text{avg}}$ on different architectures and datasets.}
\label{tab:accuracy}
\vspace{0.1em}
\begin{tabular}{c c|c c c}
\toprule
\textbf{Dataset} & \textbf{\% of training data} & \textbf{BP} & \textbf{PC}-$\mathcal{I}_{\text{fw}}$ & \textbf{PC}-$\mathcal{I}_{\text{avg}}$ \\ \midrule
\multirow{3}{*}{MNIST} & $1.0$ & $98.32\% \pm \scriptstyle 0.05\%$ & $98.26\% \pm \scriptstyle 0.15\%$ & $\mathbf{98.73\%} \pm \scriptstyle 0.07\%$ \\ 
                       & $0.5$ & $97.61\% \pm \scriptstyle 0.28\%$ & $97.84\% \pm \scriptstyle 0.29\%$ & $\mathbf{98.41\%} \pm \scriptstyle 0.08\%$ \\ 
                       & $0.25$ & $96.79\% \pm \scriptstyle 0.12\%$ & $97.09\% \pm \scriptstyle 0.09\%$ & $\mathbf{97.70\%} \pm \scriptstyle 0.09\%$ \\ \hline
\multirow{3}{*}{FashionMNIST} & $1.0$ & $89.55\% \pm \scriptstyle 0.16\%$ & $88.50\% \pm \scriptstyle 0.47\%$ & $\mathbf{89.80\%} \pm \scriptstyle 0.20\%$ \\ 
                              & $0.5$ & $88.09\% \pm \scriptstyle 0.10\%$ & $87.81\% \pm \scriptstyle 0.26\%$ & $\mathbf{88.88\%} \pm \scriptstyle 0.13\%$ \\
                              & $0.25$ & $86.58\% \pm \scriptstyle 0.20\%$ & $86.55\% \pm \scriptstyle 0.57\%$ & $\mathbf{87.64\%} \pm \scriptstyle 0.17\%$ \\ \hline
\multirow{3}{*}{KMNIST} & $1.0$ & $92.38\% \pm \scriptstyle 0.08\%$ & $92.13\% \pm \scriptstyle 0.38\%$ & $\mathbf{93.37\%} \pm \scriptstyle 0.29\%$ \\ 
                        & $0.5$ & $89.86\% \pm \scriptstyle 0.52\%$ & $90.13\% \pm \scriptstyle 0.33\%$ & $\mathbf{91.87\%} \pm \scriptstyle 0.20\%$ \\ 
                        & $0.25$ & $87.18\% \pm \scriptstyle 0.20\%$ & $87.66\% \pm \scriptstyle 0.24\%$ & $\mathbf{89.45\%} \pm \scriptstyle 0.29\%$ \\ \hline \hline
\multirow{3}{*}{CIFAR10} & $1.0$ & $\mathbf{75.82\%} \pm \scriptstyle 0.19\%$ & $66.72\% \pm \scriptstyle 1.53\%$ & $75.41\% \pm \scriptstyle 0.39\%$ \\ 
                        & $0.5$ & $70.67\% \pm \scriptstyle 0.28\%$ & $64.25\% \pm \scriptstyle 1.92\%$ & $\mathbf{71.37\%} \pm \scriptstyle 0.65\%$ \\ 
                        & $0.25$ & $64.41\% \pm \scriptstyle 0.44\%$ & $63.83\% \pm \scriptstyle 0.77\%$ & $\mathbf{66.13\%} \pm \scriptstyle 0.28\%$ \\     
\bottomrule
\end{tabular}
\end{table*}
\begin{figure*}[t]
    \centering
    \begin{subfigure}{.59\textwidth}
        \includegraphics[width=\columnwidth]{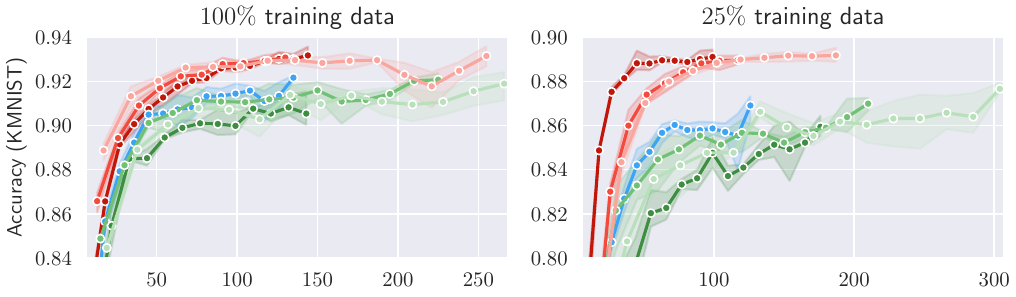}
        \includegraphics[width=\columnwidth]{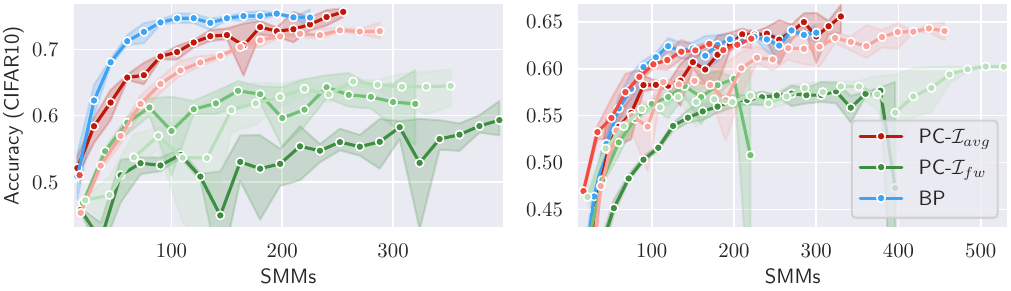}
    \end{subfigure}
    \hfill
    \begin{subfigure}{.39\columnwidth}
        \includegraphics[width=\columnwidth]{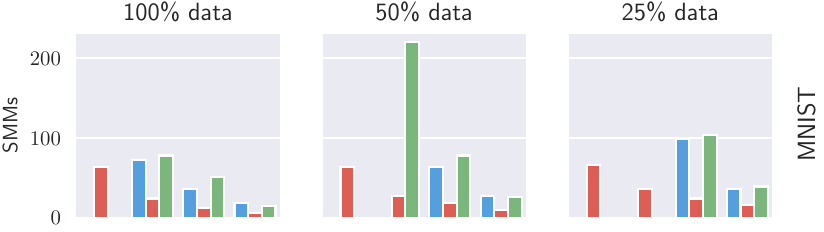}
        \includegraphics[width=\columnwidth]{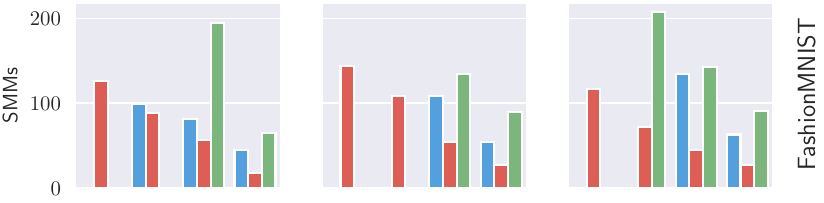}
        \includegraphics[width=\columnwidth]{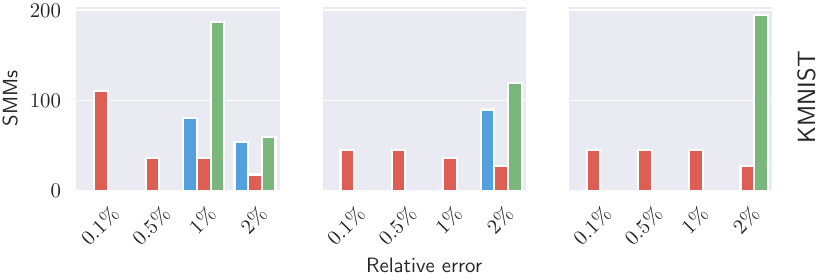}
    \end{subfigure}%
    \caption{Performance of the different learning algorithms. (Left) Learning curves representing test accuracy over epochs, scaled by the amount of SMMs. KMNIST was the dataset for which PC-$\mathcal{I}_{\text{avg}}$ performed the best relative to BP, while CIFAR10 was the worst. Shades of color indicate different $T$ values (darker $=$ lower; KMNIST: $T \in \{3,5,7\}$, CIFAR10: $T \in \{5,6,7\}$). For clarity, only the best performing $T$ values are displayed. Each curve is truncated at its highest accuracy value. (Right) Training SMMs required to achieve a given relative error (e.g., $1\%$ on KMNIST means reaching an accuracy $a \geq 93.37 \times 0.99 = 92.44\%$). PC-$\mathcal{I}_{\text{avg}}$ is the best performing method on all datasets and data splits, reaching higher accuracy and being significantly faster. Bars are missing where the target error was not reached by a method.}
    \label{fig:accuracy_over_SMMs}
\end{figure*}

\subsection{Supervised learning}
We trained the same 5-layers fully connected network
for classification on the MNIST \cite{MNIST, lecun1998gradient}, FashionMNIST \cite{FashionMNIST}, and KMNIST \cite{KMNIST} datasets. We evaluated the highest test accuracy achieved by each training method, as well as their efficiency, measured as improvements over epochs (and corresponding SMMs). We perform the same benchmarks using AlexNet on CIFAR10 \cite{CIFAR10}. 

\textbf{Setup:} We perform an exhaustive grid hyperparameter search and train each model on multiple seeds to ensure the statistical relevance of our results. The full details of the experiment are given in the supplementary material.
We observed that a hybrid initialization that combines $\mathcal{I}_{\text{fw}}$ for the first $m$ hidden layers and $\mathcal{I}_{\text{avg}}$ for the remaining $L - 1 - m$ ones performs the best. Thus, we add $m$ to the hyperparameter search. Empirically $m \approx \frac{L}{2} + 1$ seems to provide the best results. Formally, for $\mathcal{I}_{\text{avg}}$, we have that
\begin{equation}
    h_{l,i}^{(0,b+1)} = 
    \begin{cases}
        \mu_l^{(0, b+1)},& \text{if } l \leq m\\
        \frac{C}{n} \sum_{\{j: y_j = y_i\}} h^{(T, b)}_{l, j} & \text{otherwise.}
    \end{cases}
\end{equation}
As computational efficiency is our focus, we constrained the hyperparameter search to small numbers of inference steps, such that $T \approx L$ and, thus, $\operatorname{SMM}_{PC} \approx \operatorname{SMM}_{BP}$, according to the equations given in \cite{salvatori2022incremental}. To keep the number of weight updates constant for all training algorithms, we fix the batch size to $n = 200$.

\textbf{Results:} According to Table~\ref{tab:accuracy}, PC-$\mathcal{I}_{\text{avg}}$ is a notable and consistent improvement over the PC-$\mathcal{I}_{\text{fw}}$ baseline across every benchmark, scoring an almost 10\% higher accuracy on the more complex AlexNet architecture on the CIFAR10 dataset and further bridging the gap between PC (and thus biologically inspired networks) and BP. In fact, PC-$\mathcal{I}_{\text{avg}}$ achieves the highest accuracy on all the other considered benchmarks. Several works reported that PC is better suited than BP for low-parameters low-data training settings \cite{salvatori2022incremental, song2024inferring}. Hence, we test each dataset using different fractions of the training data and confirm that PC-$\mathcal{I}_{\text{avg}}$ behaves analogously, with the performance gap with BP growing the smaller the training set. Fig.~\ref{fig:accuracy_over_SMMs} (left) shows the training efficiency of PC-$\mathcal{I}_{\text{avg}}$ which, not only reaches a better accuracy, but requires fewer SMMs than PC-$\mathcal{I}_{\text{fw}}$ to do so. Except that with CIFAR10 ($100\%$ data), this is also true in relation to BP. Finally, we study how quickly each training algorithm can reach any given accuracy score. As reported in Fig.~\ref{fig:accuracy_over_SMMs} (right), PC-$\mathcal{I}_{\text{avg}}$ outperforms both BP and PC-$\mathcal{I}_{\text{fw}}$, requiring up to an order of magnitude fewer SMMs than the latter (e.g., MNIST, $50\%$ training data, $0.5\%$ relative error). Our findings reveal the ability to train PCNs with a reduced inference step count, $T < L$. PC-$\mathcal{I}_{\text{avg}}$ maintains consistent performance even at lower values of $T$ in contrast to PC-$\mathcal{I}_{\text{fw}}$, allowing for comparable performance at a lower computational expense~(Fig.~\ref{fig:accuracy_over_T}).

\subsection{Unsupervised learning}
We train a 4-layers generative decoder-only architecture on the FashionMNIST dataset to perform image generation. Under this settings, we have access only to the output of the model (i.e., the target image). As per Proposition \ref{thm:forward_dag}, $I_{\text{fw}}$ cannot be used in such instances, thus we compare $I_{\text{mem}}$ with the remaining initialization techniques. Analogously, during evaluation, we cannot perform a forward pass to compute the network's output, so we also compare inference speed of the different methods.

\textbf{Setup:} We perform an exhaustive hyperparameter search and report the best results for each method and $T$ value. More details about the exact experimental setup are given in the supplementary material. We compare the reconstruction loss at several inference step counts for both training, $T_{\text{train}}$, and evaluation, $T_{\text{eval}}$.

\begin{figure}[t]
    \centering
    \includegraphics[width=0.55\columnwidth]{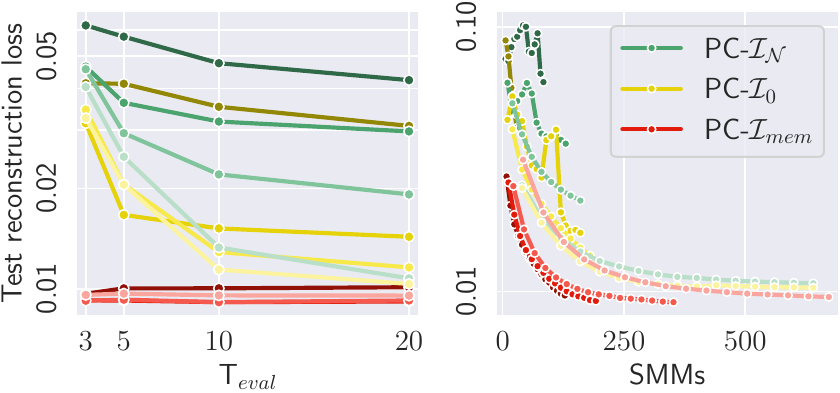}
    \caption{Reconstruction loss for different initializations on the FashionMNIST dataset. (Left) Best reconstruction loss achieved for each combination of $T_{\text{eval}}$ and $T_{\text{train}}$. Shades of color indicate different $T_{\text{train}}$ values. (Right) Learning curves of each method. For clarity, only models with $T_{\text{eval}} = T_{\text{train}} \in \{3, 5, 10, 20\}$ (darker $=$ lower) are displayed. Each curve is truncated when reaching its best reconstruction loss.}
    \label{fig:analysis_unsupervised}
\end{figure}
\begin{figure}[t]
    \centering
    \includegraphics[width=0.45\columnwidth]{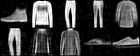}%
    \hspace{1em}%
    \includegraphics[width=0.45\columnwidth]{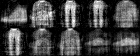}
    \caption{Reconstruction examples with $T_{\text{eval}} = T_{\text{train}} = 3$ for PC-$\mathcal{I}_{\text{mem}}$ (left) and PC-$\mathcal{I}_{0}$ (right). The difference in image quality is evident between the two methods. Examples for other $T$ values are provided in the supplementary material.}
    \label{fig:reconstructions_unsupervised}
\end{figure}

\textbf{Results:} PC-$\mathcal{I}_{\text{mem}}$ is the best performing method, producing high quality reconstructions even at low $T$ values for both training and evaluation (Fig.~\ref{fig:reconstructions_unsupervised}), resulting in clear and detailed images. The same cannot be said for the other methods. According to Fig.~\ref{fig:analysis_unsupervised} (left), which reports the best achieved reconstruction loss by each model for every combination of $T_{\text{train}}$ and $T_{\text{eval}}$ hyperparameters, the performance of PC-$\mathcal{I}_{\text{mem}}$ is unaffected by the choice for both values. Instead, PC-$\mathcal{I}_{\mathcal{N}}$ and PC-$\mathcal{I}_{0}$ show improvements even at the highest values tested ($T_{\text{train}} = 20$ and $T_{\text{eval}} = 20$), which translates to almost an order of magnitude more inference steps (i.e., $3$ against $20$), and thus SMMs, to reach the same image quality, both in training and evaluation. This is also reflected by the learning curve of each method (Fig.~\ref{fig:analysis_unsupervised} (right)). We note that PC-$\mathcal{I}_{\text{mem}}$ is the most stable algorithm, with smooth convergence even at low $T$ values.

\section{Related Work}
\label{related-work}
In the past few years, it has been shown that PC can approximate BP gradient updates \cite{whittington2017approximation, millidge2022approximates} and even be algorithmically equivalent to it \cite{salvatori2022reverse} under specific constraints. It has been proven that PC generates weight updates that capture higher-order information of the loss landscape \cite{innocenti2023understanding, alonso2023understanding}, and, ultimately, are better aligned with the task target which guarantees faster convergence. In later years, most of the focus has been into scaling up the performance of such models, and making training feasible on deeper models such as ResNets \cite{qi2025towards, goemaere2025error}. However, on GPUs, BP is still the preferable option due to its lower computational cost.


In this work, continuous Hopfield networks \cite{ramsauer2020hopfield} have been chosen to implement associative memories due to their straightforward integration and low computational requirements. However, it has been shown that PCNs can also be used as associative memories which resemble the internal working of the hippocampus in the human brain \cite{salvatori2021associative, yoo2022bayespcn, tang2023recurrent}. Despite their current computational cost, they could provide an alternative that better integrates into  the PC framework and scales to more complex models for which more robust memorization capabilities may be necessary.

Different works are pursuing the long-standing goal of studying biologically plausible alternative to BP, due to the inherent \emph{locality} of their operations and synaptic updates, useful in the context of neuromorphic hardware \cite{scellier2023energy,hagiwara2024design,ororbia2023spiking}.
This attribute, when applied within hardware paradigms, facilitates the extensive parallelization of computational tasks, leading to reductions in latency and power consumption, often with no supervision signals. This stands in stark contrast to the conventional Von-Neumann architectures, where the distinct segregation of memory and processing units imposes considerable delays and computational demands \cite{khacef2023spike}.
Other interesting alternatives to BP, also promising from the perspective of neuromorphic computing, are the aforementioned equilibrium propagation \cite{scellier2017equilibrium}, SoftHebb \cite{moraitis2022softhebb,journe2022hebbian}, and direct feedback alignment \cite{nokland2016direct}.

\section{Conclusion}
\label{conclusion}
In this work, we have tackled the problem of efficiency in predictive coding models from the perspective of reducing the number of iterations needed to perform a meaningful update of the weight parameters. To this end, we have studied how adding an external memory at each level of the hierarchy, allows us to use data information to better initialize the latent variables. In classification tasks, we have simply memorized the average value of the neurons at convergence for every label, and use it as initialization. For generation tasks, we have used Hopfield layers \cite{ramsauer2020hopfield} as external memories. We have also shown, theoretically and empirically, that our method not only improves the efficiency, but also leads to a better performance in terms of both test accuracies and reconstruction loss. An interesting future direction would be to generalize our results to other energy-based learning methods, such as equilibrium propagation, as well as extending this method to more complex tasks already tackled in the predictive coding literature, such as graph neural networks \cite{byiringiro2022robust} and transformer models \cite{pinchetti2022beyond}.

\bibliographystyle{acm}
\bibliography{refs}

\newpage
\
\appendix

\section{Notation.}
We refer to a layer $h$ (i.e., a cluster of neurons) in a PCN $\mathcal{M}$ as the following vector:
\begin{equation}
    h_{l,i}^{(t,b)},
\end{equation}
where the indices used are explained in Tab.~\ref{tab:notation}.
\begin{table}[h!]
    \centering
    \caption{Notation used to identify a layer $h$.}
    \begin{tabular}{c|l|l}
        \toprule
        \textbf{Symbol} & \textbf{Value range} & \textbf{Meaning} \\
        \midrule
        $t$ & $0, ..., T$ & PC inference step counter \\
        $b$ & - & Training batch counter \\
        $l$ & $0, ..., L$ & Layer index \\
        $i$ & $1, ..., n$ & Sample index in batch \\
        \bottomrule
    \end{tabular}
    \label{tab:notation}
\end{table}

\section{Proof of Proposition \ref{thm:forward_dag}}

\begin{figure}[t]
    \centering
    \includegraphics[width=0.7\columnwidth]{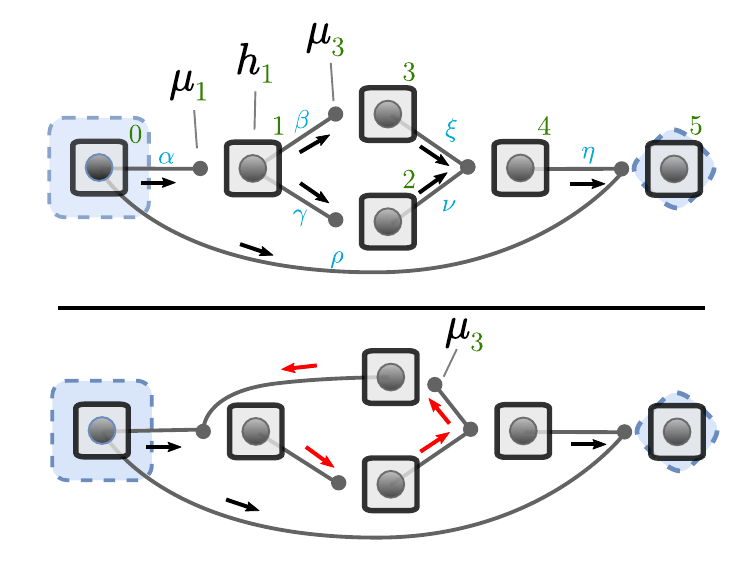}
    \caption{A concrete example of two PCNs with differing flow direction. Both PCNs have the same underlying undirected graph (node numbers -in green- and weights -in blue- are indicated only on the top graph, but the same holds for the bottom example of a graph). The square, light blue neuron denotes the input neuron. Nodes are numbered in green (only shown for the top PCN). (Top) A PCN represented by a DAG in which computation proceeds in a forward direction. This is similar to the PCN from Figure~\ref{fig:pcn}, except that this PCN does not have a layered structure, but is more general. (Bottom) A PCN with a different edge orientation, with a loop (in red). This PCN falls outside the category of DAGs.}
    \label{fig:pcn with loops}
\end{figure}

Consider  the bottom PCN as in Figure~\ref{fig:pcn with loops}. We proceed by proving the (logically equivalent) contrapositive statement: If the PCN is not a DAG, then a forward initialization cannot exist. For this, consider the bottom graph from Figure~\ref{fig:pcn with loops}, which represents a PCN that is not defined on a DAG, but a more general graph, since it has a loop. Now, we proceed by contradiction and assume a forward initialization exists. Let the neurons be numbered $0,\ldots,5$ and the edge weights be denoted by $\alpha, \beta, \gamma, \mu, \nu, \eta \in\mathbf{R}$. If the input neuron ($0$) is initialized with $s\in \mathbb{R}$, and $f$ is some activation function, then $\mu_{1}=f(\alpha s)$. We  thus have, by the definition of forward initialization $\mu_1 = h_1 = f(\alpha s)$. We now enter the loop part of the graph. It follows that $\mu_2 = h_2 = f(\gamma f (\alpha s))$ and $\mu_3 = h_3 = f(\xi f(\gamma f (\alpha s)))$. By the definition of forward initialization, we then obtain $h_1 = f( \beta f (\xi f(\gamma f (\alpha f(s)))))$. But, in general (unless $f$ and the weights have trivial values, this will be different from $h_1 = f(\alpha s)$. So, our assumption was wrong, and forward initialization cannot exist, which concludes the proof.

We remark that the converse implication does not hold: If a PCN is a DAG, then forward initialization may or may not exist, depending on the choice of input neurons.

We highlight an example where a forward initialization exists: Consider the top graph from Figure~\ref{fig:pcn with loops}, with neurons and weight numbered in the same way as in the proof above. Again we have, if the input neuron ($0$) is initialized with $s\in \mathbb{R}$, and $f$ is some activation function, that $\mu_{1}=f(\alpha s)$. By definition of forward initialization, we also set $h_1 = f(\alpha s)$. Continuing in this manner, we obtain $\mu_2=h_2 = f(\gamma f(\alpha s))$, $\mu_3=h_3 = f(\beta f(\alpha s))$,  $\mu_4 = h_4 = f(\nu f( \gamma f(\alpha s)) + \xi f( \beta f(\alpha s)) + \rho s) $, and, finally $$\mu_5 = h_5 = f( \eta (f(\nu f( \gamma f(\alpha s)) + \xi f( \beta f(\alpha s)) + \rho s))).$$
This shows that a forward initialization, where at each neuron $h_i$ and $\mu_i$ agree, exists.

Now consider the case where the input $s$ is mapped to node $1$. In this case, no edge points to node $0$, so there is no trivial possibility to find $h_0$ such that $\mu_1 = h_1$ and $\mu_4 = h_4$. Given that edge may represent transformation that do not have an inverse, such value for $h_0$ may even not exist in the general case. This case is trivial in a sense, but nonetheless, such examples form an important subset of PCN architectures.

\begin{algorithm}[t]
    \caption{Learning on a mini-batch $b$ with data $O = [(x, y)_i]_{i \in \{1,\dots,n\}}$ via PC-$\mathcal{I}_{\text{avg}}$.}
    \label{alg:PC_avg}
    \begin{algorithmic}[1]
    \REQUIRE $h_0,i^{(t, b)} = x_i$, $h_L,i^{(t, b)} = y_i$, \\ $\forall~t~\in~\{0, \dots, T\}$.

    \COMMENT{Neurons initialization}
    \FOR{$l=1$ to $L-1$}
        \STATE initialize $h^{(0, b)}_{l,i}$ according to Eq.~\eqref{eq:avg_init}
    \ENDFOR

    \COMMENT{Inference phase}
    \FOR{$t=1$ to $T$}
        \FOR{each layer $l$}
            \STATE update $h_{l,i}^{(t)}$ to minimize $\mathcal{F}$ via Eq.~\eqref{eq:delta-x}
        \ENDFOR
    \ENDFOR

    \COMMENT{Learning phase}
    \FOR{each layer $l \in {0,\dots,L-1}$}
        \STATE update $W_{l}^{(b)}$ to minimize $\mathcal{F}$ via Eq.~\eqref{eq:delta-w}
    \ENDFOR
    \end{algorithmic}
\end{algorithm}
\begin{algorithm}[t]
    \caption{Learning on a mini-batch $b$ with data $O~=~[(x, y)_i]_{i \in \{1,\dots,n\}}$ via PC-$\mathcal{I}_{\text{mem}}$.}
    \label{alg:PC_mem}
    \begin{algorithmic}[1]
    \REQUIRE $h_0,i^{(t, b)} = x_i$, $h_L,i^{(t, b)} = y_i$, $\forall~t~\in~\{0, \dots, T\}$; $V_l$ initialised as $V_l = A_l^+ h^{(0)}_{\mathcal{I}_{\text{fw}},l}$.

    \COMMENT{Neurons initialization}
    \FOR{$l=1$ to $L-1$}
        \STATE initialize $h^{(0, b)}_{l,i}$ according to Eq.~\eqref{eq:mem_init}
    \ENDFOR

    \COMMENT{Inference phase}
    \FOR{$t=1$ to $T$}
        \FOR{each layer $l \in {1,\dots,L-1}$}
            \STATE update $h_{l,i}^{(t)}$ to minimize $\mathcal{F}$ via Eq.~\eqref{eq:delta-x}
        \ENDFOR
    \ENDFOR

    \COMMENT{Learning phase}
    \FOR{each layer $l \in {0,\dots,L-1}$}
        \STATE update $W_{l}^{(b)}$ to minimize $\mathcal{F}$ via Eq.~\eqref{eq:delta-w}
        \STATE update $Q_l, b_l, K_l, V_l$ to minimize $\mathcal{L}_\mathcal{H}$ (Eq.~\eqref{eq:mem_loss})
    \ENDFOR
    \end{algorithmic}
\end{algorithm}
\section{Implementation Details}
We report the hyperparameters used and the experimental methodology followed for each experiment presented in the main body. Pseudocode for PC-$\mathbf{I}_{\text{avg}}$ and PC-$\mathbf{I}_{\text{mem}}$ is provided in Alg.~\ref{alg:PC_avg} and Alg.~\ref{alg:PC_mem} respectively. For all neural networks considered, the weights were initialised with the default Kaiming unifrom initialisation as, to the best of our knowledge, there are no results suggesting different initialisations for PC networks.

\subsection{Image classification}

\textbf{Fully connected neural network:} We trained a 5-layers multi-layer perceptron with hidden width $d = 512$. We used batches of size of $n = 200$ and trained for $e = 16$ epochs and $s = 3$ seeds. We used \textit{AdamW} as optimizer for the transformation weights, and $SGD$ for the neurons. We performed the following hyperparameter search and picked the best (averaged over seeds) test accuracy achieved by each method on each training data fraction. In Fig.~\ref{fig:training_comparison} (left), we report the best training curves for each $T$ value tested. In Fig.~\ref{fig:training_comparison} (right), for each relative test error value, we reported the lowest number of SMMs required to achieve such value among all the trained model for each training method.

\begin{table}[t]
    \centering
    \begin{tabular}{c|c c}
    \toprule
         Dataset & PC-$\mathcal{I}_f$ & PC-$\mathcal{I}_{\text{avg}}$\\
    \midrule
         MNIST & 4.71s & 1.38s \\
         KMNIST & 2.82s & 0.66s \\
         FashionMNIST & 4.42s & 2.19s \\
    \bottomrule
    \end{tabular}
    \caption{Required training time on different datasets ($100\%$ of the training data).}
    \label{tab:running_time}
\end{table}
All the models are trained for $e_{MLP} = 16$ and $e_{AlexNet} = 24$ epochs across all the experiments. 
Nonetheless, when comparing our test results with other benchmarks in the literature, we find that we achieve similar or better performance to converged models (e.g., \cite{salvatori2022incremental}). To give an idea on how SMMs translates to actual running time, in Tab.~\ref{tab:running_time}, we report the training time of PC-$\mathcal{I}_f$ and PC-$\mathcal{I}_{\text{avg}}$ to achieve the best shared accuracy (i.e., the same relative error as reported by Fig.~\ref{fig:accuracy_over_SMMs} (right)). Results are only indicative, as the framework used does not allow for parallel execution of the layers in the PCN. To avoid any overhead, training data was preloaded on the GPU.

\textit{Hyperparameters:}
\begin{itemize}
    \item training data fraction $\in \{0.25, 0.5, 1.0\}$;
    \item $\beta \in \{0.0001, 0.0003, 0.0005, 0.001, 0.003, 0.005\}$;
    \item activation function $\in \{\text{gelu}, \text{elu}, \text{tanh}, \text{leaky\_relu}, \text{relu}\}$;
\end{itemize}
(only for PC)
\begin{itemize}
    \item $\alpha \in \{0.003, 0.005, 0.01, 0.03, 0.05, 0.1, 0.3, 0.5\}$;
    \item $T$ $\in \{2,3,4,5,6,7\}$;
\end{itemize}
(only for PC-$\mathcal{I}_{\text{avg}}$)
\begin{itemize}
    \item $m \in \{1,2,3\}$.
\end{itemize}
For Fig.~\ref{img:init_summary}, we repeat the same experiment with the remaining initialization methods and with $T~\in~\{5, 10, 25, 50, 100\}$.

\textbf{AlexNet neural network:} the training setup remains unchanged, but we perform the following hyperparameter search.

\begin{figure*}[t]
    \centering
    \includegraphics[width=1\linewidth]{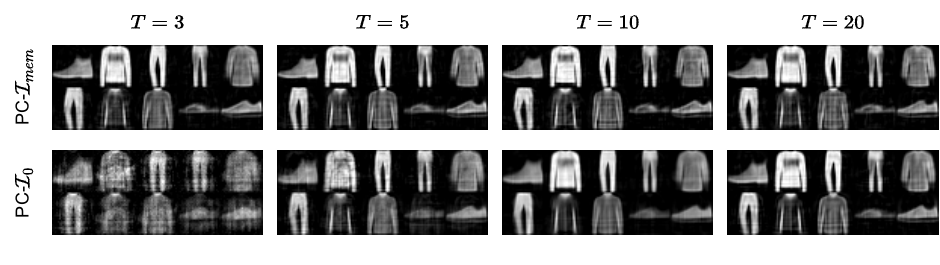}
    \caption{Examples of reconstructed images at different inference steps $T$ counts.}
    \label{fig:recs_full}
\end{figure*}

\textit{Hyperparameters:}
\begin{itemize}
    \item training data fraction $\in \{0.25, 0.5, 1.0\}$;
    \item $\beta \in \{0.0003, 0.0005, 0.001, 0.003\}$;
    \item activation function $\in \{\text{gelu}, \text{elu}, \text{tanh}, \text{leaky\_relu}, \text{relu}\}$;
\end{itemize}
(only for PC)
\begin{itemize}
    \item $\alpha \in \{0.01, 0.03, 0.05, 0.1\}$;
    \item $T$ $\in \{4,5,6,7,8,9\}$;
\end{itemize}
(only for PC-$\mathcal{I}_{\text{avg}}$)
\begin{itemize}
    \item $m \in \{4,5,6\}$.
\end{itemize}

Note that also PC-$\mathbf{I}_{\text{mem}}$ was tested on classification tasks using the best hyperparameters found for PC-$\mathbf{I}_{\text{avg}}$ using $p_\mathcal{H} \in \{8,16,24,32\}$ memory patterns per layer. No significant difference was observed compared to PC-$\mathbf{I}_{\text{avg}}$, with slight variations based on the value of $p_\mathcal{H}$. Thus, when labels are available, PC-$\mathbf{I}_{\text{avg}}$ is the preferred initialisation method as it does not introduce any memory overhead required by the \textit{HopfieldLayers}.

\subsection{Image reconstruction}
We trained a 4-layers multi-layer perceptron with hidden width $d=256$ to act as a decoder only architecture with a $d_0 = 64$ wide bottleneck layer $h_0$. Similarly to above, we used batches of size of $n = 200$ and trained for $e = 16$ epochs and $s = 3$ seeds. We used \textit{AdamW} as optimizer for the transformation weights, and $SGD$ for the neurons.
During training, we fix the output layer $h_L$ to a given image and minimise the energy according to Eqs.~\eqref{eq:delta-x} and~\eqref{eq:delta-w}. During evaluation, we test the ability of the network to represent and reconstruct a given test image by fixing it to the output layer, minimising the energy by acting exclusively on the neurons during inference, and finally forwarding the value of layer $h_0$ through the network to produce an output image. Example of the best reconstructions for each $T$ are shown in Fig.~\ref{fig:recs_full}. Similar results were achieved on the MNIST and KMNIST datasets. \\
We performed the following hyperparameter search and picked the best (averaged over seeds) test reconstruction error achieved by each method. Since we need to perform inference even during evaluation, we have a separate set of hyperparameters for it.

\textit{Hyperparameters:}
\begin{itemize}
    \item activation function $\in \{\text{gelu}, \text{tanh}, \text{leaky\_relu}\}$;
    \item $\beta \in \{0.0001, 0.0003, 0.0005\}$;
    \item $\alpha_{\text{train}} \in \{0.05, 0.1, 0.5\}$;
    \item $\alpha_{\text{eval}} \in \{0.05, 0.1, 0.5\}$;
    \item $T_{\text{train}}$ $\in \{3, 5, 10, 20\}$;
    \item $T_{\text{eval}}$ $\in \{3, 5, 10, 20\}$.
\end{itemize}
For PC-$\mathcal{I}_{\text{mem}}$, we use a single Hopfield network $\mathcal{H}$ for the input layer $h_0$, and forward its value through the remaining hidden neurons to initialize the network. We store $p_\mathcal{H} = 24$ patterns in memory, and use a $a_\mathcal{H} = 16$ heads multi-attention retrieval mechanism, with an embedded size of $e_\mathcal{H} = 128$. We searched over $\delta \in \{1.0, 10.0, 50.0, 200.0, 500.0\}$ and found that $\delta = 200.0$ seems to provide the best results, although the difference is negligible.

\end{document}